\documentclass[journal]{IEEEtran}

\usepackage{booktabs}
\usepackage{multirow}
\usepackage{graphicx}
\usepackage{amsmath}
\usepackage{amssymb}
\usepackage{cite}
\usepackage{url}
\usepackage{hyperref}
\usepackage{verbatim}
\usepackage{balance}

\begin{document}

\title{Meta-CLIP: Metadata-Conditioned Contrastive Learning for Zero-Shot Disease Inference}
\author{Haozhe Luo, Nirosh Sivanesan, Shelley Zixin Shu, Francesco Nitti, and Mauricio Reyes%
\thanks{Haozhe Luo is with the University of Bern, Bern, Switzerland (e-mail: haozhe.luo@unibe.ch).}%
\thanks{Nirosh Sivanesan, Shelley Zixin Shu, Francesco Nitti, and Mauricio Reyes are with the ARTORG Center for Biomedical Engineering Research, University of Bern, Bern, Switzerland (e-mail: Sivanirosh@gmail.com; zixin.shu@unibe.ch; francesco.nitti@unibe.ch; mauricio.reyes@unibe.ch).}}

\maketitle

\begin{abstract}
Medical vision--language pretraining learns transferable representations from paired images and radiology reports, yet differences in scanners and acquisition protocols can impair transfer. Existing approaches primarily enrich clinical semantics, leaving structured acquisition information in DICOM headers underused. We introduce Meta-CLIP, a metadata-conditioned contrastive framework that uses acquisition similarity to reweight cross-modal negatives during pretraining. A kernel over numerical and categorical DICOM fields defines a bounded pairwise margin. This margin increases the contribution of acquisition-similar negatives, emphasizing distinctions among studies that share imaging conditions. The kernel remains fixed during image--text optimization, and inference follows the standard dual-encoder architecture without requiring metadata. We evaluate Meta-CLIP on chest radiography and volumetric chest CT through zero-shot classification, appearance perturbations, grounding, and representation analyses. On CT-RATE, Meta-CLIP improves macro-AUROC over CT-CLIP by 1.41 points, with a 95\% paired bootstrap confidence interval of 0.96--1.85 points. It also achieves the highest macro-AUROC on four of five CXR benchmarks. An exploratory BrgSA evaluation extends the analysis to another CT backbone. These findings support acquisition metadata as a practical training-time prior for medical vision--language learning. Code will be made publicly available upon publication.
\end{abstract}


\section{Introduction}
Vision--language pretraining (VLP) offers a scalable way to learn medical
image representations from routinely collected radiology reports. Aligning
images and text in a shared embedding space enables zero-shot disease
classification and cross-modal retrieval, reducing the need for task-specific
annotations
\cite{radford2021learning,zhang2022contrastive,huang2021gloria,boecking2022making}.
Progress in this area has strengthened clinical supervision through biomedical
text representations, medical knowledge, and large-scale volumetric
pretraining \cite{zhang2023knowledge,wu2023medklip,hamamci2026generalist}.

Transfer across acquisition settings presents a complementary challenge.
Scanner vendors, exposure settings, and reconstruction protocols can change
image appearance without changing the underlying pathology. Medical
vision--language models remain sensitive to imaging artifacts and distribution
shifts \cite{cheng2025understanding}. In standard contrastive learning,
unmatched image--report pairs contribute according to their embedding
similarity. The objective has no explicit acquisition prior to guide which
negative comparisons it emphasizes.

DICOM headers make this context available through structured acquisition
fields. Metadata has supported representation learning in ultrasound, retinal
imaging, and longitudinal OCT
\cite{hu2020self,holland2024metadata,lee2025preti}. For image--report
pretraining, our insight is that acquisition cues shared by two studies are
less informative for determining which report corresponds to which image.
Emphasizing comparisons between acquisition-similar studies encourages
alignment to rely on other image--report evidence. We use acquisition
similarity as a prior for weighting these contrastive negatives.

\begin{figure*}[!t]
    \centering
    \includegraphics[width=0.92\textwidth]{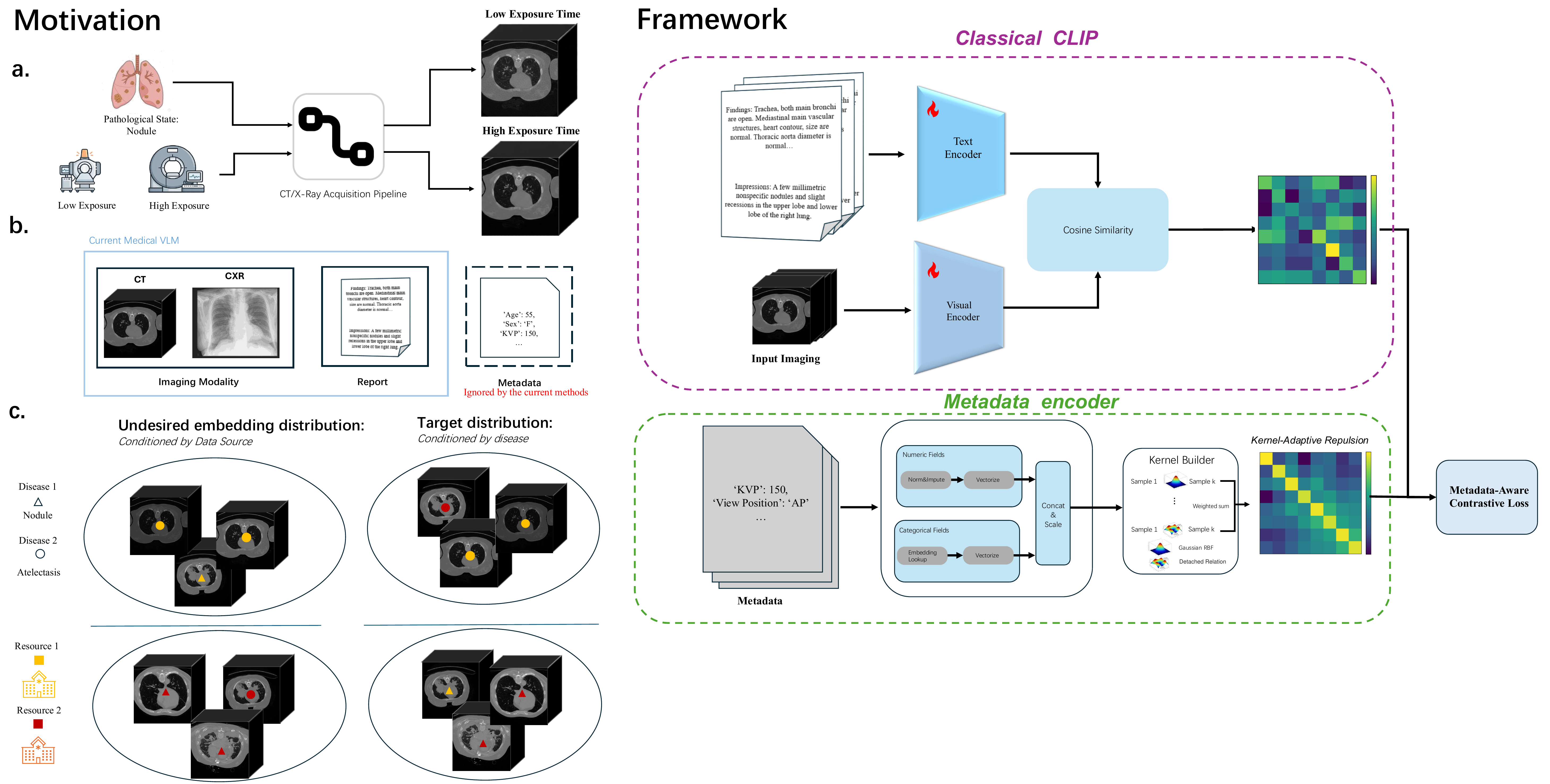}
    \caption{Motivation and overview of Meta-CLIP. Left: (a) Acquisition protocols and scanner vendors influence image appearance. (b) Standard image--report pretraining leaves structured acquisition metadata unused. (c) Acquisition variation can shape the learned representation. Right: Meta-CLIP constructs a pairwise acquisition kernel from numerical and categorical DICOM fields and uses it to modulate negative logits in the contrastive objective. Metadata is used during pretraining and is unnecessary at inference.}
    \label{fig:overview}
\end{figure*}

We introduce \emph{Meta-CLIP}, which implements this prior through
metadata-conditioned margins for chest radiography and volumetric chest CT
(Fig.~\ref{fig:overview}). A kernel combines numerical acquisition fields and
categorical scanner attributes into a pairwise similarity. This similarity
defines bounded margins that raise negative logits in both contrastive
directions, giving acquisition-similar negatives greater weight while leaving
positive logits unchanged. The kernel stays fixed during image--text
optimization. Inference retains the standard dual-encoder architecture and
requires no metadata.

Experiments cover CXR and CT classification, external transfer, and analyses
of robustness, grounding, and representation structure. Meta-CLIP improves
CT-RATE macro-AUROC over CT-CLIP by 1.41 points and achieves the highest
macro-AUROC on four of five CXR benchmarks.

The contributions of this work are threefold:
\begin{itemize}
    \item We formulate acquisition similarity as a prior over cross-modal
    negatives and construct this relation from numerical and categorical
    DICOM fields.
    \item We develop a symmetric contrastive objective that emphasizes
    acquisition-similar negatives through bounded pairwise margins, preserving
    the underlying encoders and their metadata-free inference procedure.
    \item We evaluate CXR and CT transfer, characterize the selected metadata,
    and examine robustness, grounding, representation structure, and
    exploratory transfer to another CT backbone.
\end{itemize}

\section{Related Work}
\subsection{Medical Vision--Language Pretraining}
Contrastive pretraining established that natural-language supervision can
produce transferable visual representations \cite{radford2021learning}.
Medical adaptations use radiology reports as scalable supervision,
progressing from global image--text alignment in ConVIRT
\cite{zhang2022contrastive} to combined global and local alignment in GLoRIA
\cite{huang2021gloria}. BioViL strengthens biomedical text representations
\cite{boecking2022making}. Further developments introduce soft prompts
\cite{chen2023towards}, granular masked alignment \cite{huang2023enhancing}, or
anatomical consistency \cite{zhou2023learning} to improve transfer.

A complementary line of work enriches report supervision with structured
medical knowledge. KAD and MedKLIP associate visual features with disease
entities and knowledge graphs \cite{zhang2023knowledge,wu2023medklip}, while
DeViDe decomposes disease descriptions into multiple semantic facets
\cite{luo2025devide}. CT-CLIP extends image--text pretraining to volumetric
chest CT at scale \cite{hamamci2026generalist}. These methods strengthen
clinical supervision and image--text alignment. Meta-CLIP complements this
direction by using acquisition context to determine the contribution of
cross-modal negatives during pretraining.

\subsection{Robustness and Vision--Language Adaptation}
Distribution shifts remain a major obstacle for medical VLP. Controlled
evaluation has shown that common medical image artifacts can substantially
change VLM predictions \cite{cheng2025understanding}. Adaptation methods seek
to improve transfer through learned prompts \cite{zhou2022learning},
training-free feature adapters \cite{zhang2021tip}, or disease-informed
adaptation for medical images \cite{zhang2024disease}. These approaches modify
prompts, features, or model parameters for the target task. They address a
different stage of learning from the construction of contrastive relations
during source pretraining.

Meta-CLIP uses acquisition information accompanying each source image to
modify the pretraining objective before target-domain inference. The resulting
model can be evaluated directly in zero-shot settings or combined with later
adaptation when target data are available.

\subsection{Metadata-Aware Medical Representation Learning}
Metadata can supply supervision beyond the image itself. DICOM fields provide
weak supervision for ultrasound pretraining \cite{hu2020self}, while patient
identity, laterality, and temporal information define relations for contrastive
learning from longitudinal retinal OCT \cite{holland2024metadata}. PRETI combines
patient metadata with learnable metadata embeddings and retinal-aware masking
for fundus representation learning \cite{lee2025preti}. Together, these studies
show that routinely available context can improve medical image
representations through weak targets, patient relations, or sample-pair
definitions in unimodal pretraining.

Our setting differs in both the metadata and its role. We use acquisition
parameters from DICOM headers to characterize protocol and scanner similarity
in paired image--report pretraining. Rather than treating metadata as an input,
prediction target, or rule for constructing positive pairs, Meta-CLIP applies
the acquisition relation as a continuous margin on cross-modal negatives.
This preserves the standard CLIP encoders and removes the need for metadata at
inference.

\begin{figure*}[!t]
    \centering
    \includegraphics[width=0.92\textwidth]{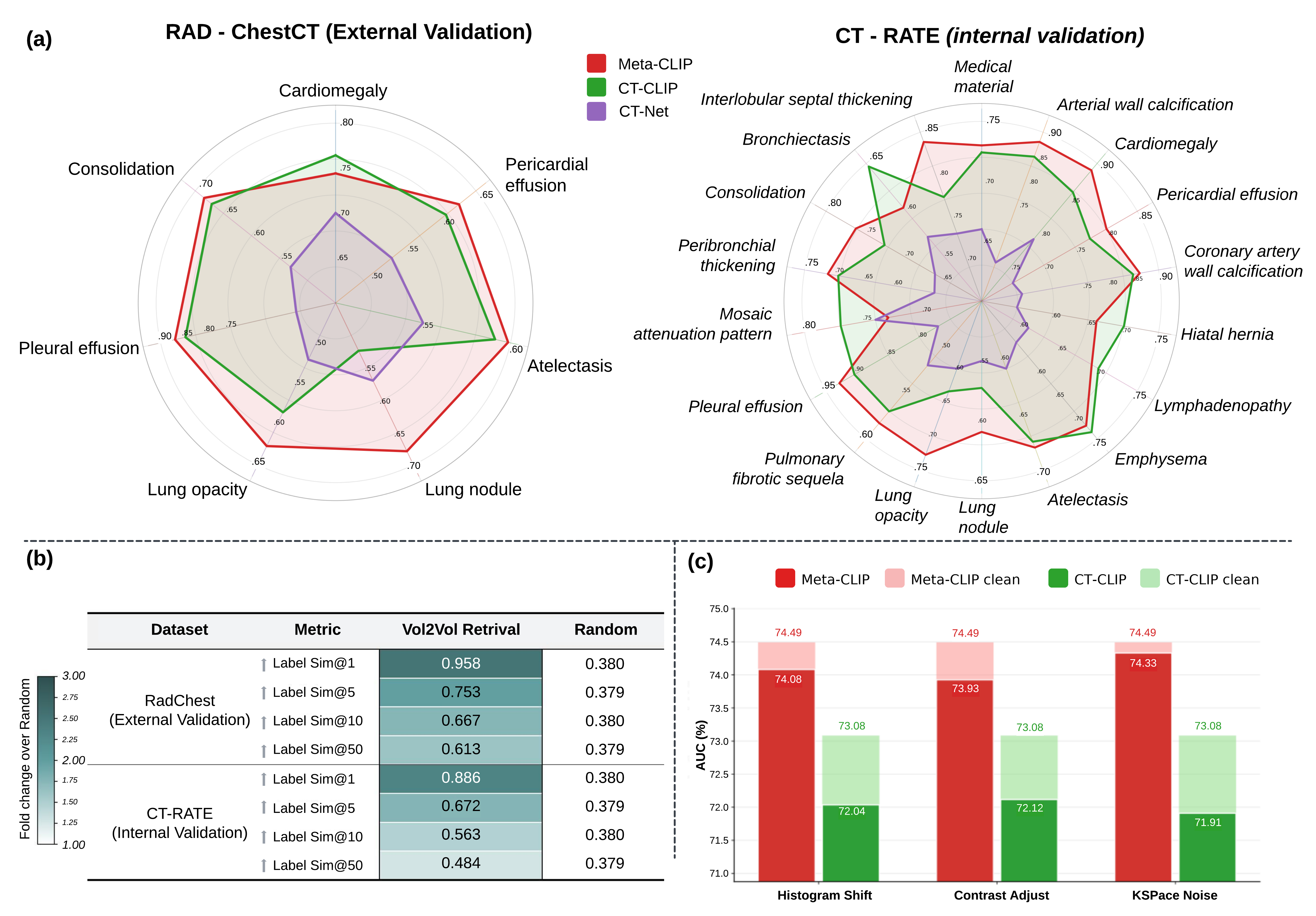}
    \caption{(a) Disease-wise AUC comparison on RAD-ChestCT and CT-RATE. (b) Volume-to-volume retrieval measured by label similarity at different top-$k$ levels. (c) Robustness under synthetic perturbations; the light bars show clean inputs and the dark bars show perturbed inputs.}
    \label{fig:ct_results}
\end{figure*}

\begin{figure*}[!t]
    \centering
    \includegraphics[width=0.90\textwidth]{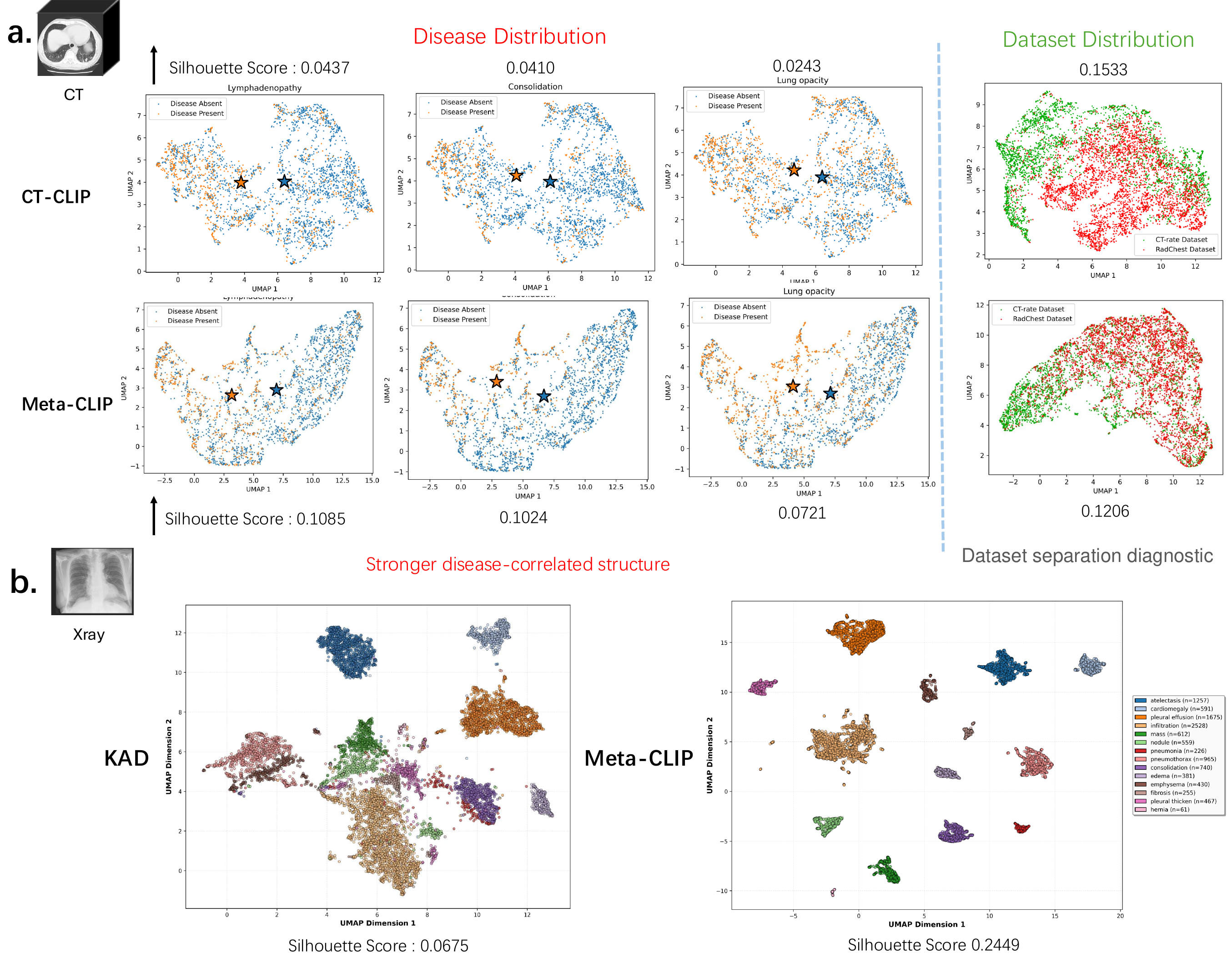}
    \caption{\textbf{Embedding structure.} UMAP visualizations and silhouette scores for (a) CT and (b) CXR. Meta-CLIP shows stronger disease-related clustering and, on CT, lower separation by dataset than CT-CLIP.}

    \label{fig:embedding_umap}
\end{figure*}

\section{Method}
Meta-CLIP introduces structured acquisition metadata into contrastive pretraining
through a pairwise margin on cross-modal negatives. As illustrated in
Fig.~\ref{fig:overview}, a kernel measures similarity between acquisition
conditions, and the resulting relation modulates logits in the contrastive
objective. The image and text encoders retain their original architecture.

\subsection{Problem Formulation}
Let $\mathcal{D}=\{(x_i,r_i,m_i)\}_{i=1}^{M}$ denote a pretraining
dataset, where $x_i$ is a CXR image or CT volume, $r_i$ is its paired report, and
$m_i$ contains structured DICOM metadata.
An image encoder $f_{\theta}$ and a text encoder $g_{\phi}$ produce
$\ell_2$-normalized representations
\begin{equation}
    z_i^{\mathrm{I}}
    = \frac{f_{\theta}(x_i)}{\|f_{\theta}(x_i)\|_2},
    \qquad
    z_i^{\mathrm{T}}
    = \frac{g_{\phi}(r_i)}{\|g_{\phi}(r_i)\|_2}.
    \label{eq:dual_embeddings}
\end{equation}
The metadata $m_i$ describes acquisition conditions such as exposure,
reconstruction, and scanner manufacturer. We use these conditions to define a
prior over unmatched pairs within each minibatch. This prior controls their
contribution to the contrastive normalizer, while matched image--report pairs
provide the positive supervision.

\subsection{Acquisition Relation Modeling}
For a minibatch $\mathcal{B}=\{(x_i,r_i,m_i)\}_{i=1}^{N}$, we construct a
pairwise acquisition kernel from numerical and categorical DICOM fields. The
two field types capture complementary aspects of image formation and scanner
identity.

\subsubsection{Numerical metadata}
Let $\mathcal{D}_{\mathrm{num}}$ index the numerical DICOM fields and let each
field be standardized independently. For a pair $(i,j)$, $\mathcal{D}_{ij}
\subseteq\mathcal{D}_{\mathrm{num}}$ contains the fields observed for both
samples, so missing values are excluded from that pairwise comparison. We
measure numerical similarity using an RBF kernel,
\begin{equation}
    K^{\mathrm{num}}_{ij}
    = \exp\!\left(
        -\sum_{d\in\mathcal{D}_{ij}}
        \frac{(m_{i,d}-m_{j,d})^2}{2\sigma_d^2}
      \right),
    \label{eq:numeric_kernel}
\end{equation}
where $\sigma_d>0$ controls sensitivity to differences in field $d$.
Standardization places fields with different units on comparable scales,
while the bandwidth determines how quickly similarity decays.

\subsubsection{Categorical metadata}
Let $\mathcal{D}_{\mathrm{cat}}$ index categorical fields such as scanner
manufacturer. Each field $c$ has a fixed lookup embedding $E_c$. With
$\widehat{E}_c(v)=E_c(v)/\|E_c(v)\|_2$, categorical similarity is
\begin{equation}
    K^{\mathrm{cat}}_{ij}
    = \frac{1}{|\mathcal{D}_{\mathrm{cat}}|}
      \sum_{c\in\mathcal{D}_{\mathrm{cat}}}
      \max\!\left(0,\left\langle
        \widehat{E}_c(m_{i,c}),
        \widehat{E}_c(m_{j,c})
      \right\rangle\right).
    \label{eq:categorical_kernel}
\end{equation}
Numerical and categorical similarities are combined through a normalized
aggregation operator:
\begin{equation}
    K^{\mathrm{meta}}_{ij}
    = \operatorname{Agg}\!\left(
      K^{\mathrm{num}}_{ij},K^{\mathrm{cat}}_{ij}
    \right).
    \label{eq:metadata_kernel}
\end{equation}
For CXR, the Gaussian relation uses a single normalized bandwidth;
for CT, per-field similarities are combined by normalized field weights.
The resulting acquisition kernel lies in $[0,1]$, with larger values denoting
more similar imaging conditions.

\subsubsection{Meta-CLIP relation}
We convert the acquisition kernel into a negative correction for every
off-diagonal pair and leave the matched image--report pair unchanged:
\begin{equation}
    K^{\mathrm{total}}_{ij}=
    \begin{cases}
      -\alpha K^{\mathrm{meta}}_{ij}, & i\ne j,\\
      0, & i=j.
    \end{cases}
    \label{eq:metadata_relation}
\end{equation}
Here $\alpha\in[0,1]$ controls the correction strength. Thus,
$K^{\mathrm{total}}_{ij}\in[-\alpha,0]$, and acquisition-similar negative
pairs receive corrections of larger magnitude. Both modalities use this
relation.

\subsection{Metadata-Conditioned Contrastive Learning}
For image-to-text retrieval, the standard CLIP similarity between image $i$ and
report $j$ is
\begin{equation}
    S^{\mathrm{I2T}}_{ij}
    = \left\langle z_i^{\mathrm{I}},z_j^{\mathrm{T}}\right\rangle.
    \label{eq:clip_similarity}
\end{equation}
The text-to-image matrix is defined analogously as
$S^{\mathrm{T2I}}_{ij}=\langle z_i^{\mathrm{T}},z_j^{\mathrm{I}}\rangle$.
We inject the relation into each off-diagonal logit using an
additive margin:
\begin{equation}
    \widetilde{S}^{q}_{ij}
    = \frac{S^{q}_{ij}
      - \mu_t\,\mathbf{1}[i\neq j]K^{\mathrm{total}}_{ij}}{\tau},
    \qquad q\in\{\mathrm{I2T},\mathrm{T2I}\},
    \label{eq:meta_logits}
\end{equation}
where $\mu_t=\mu_{\max}s(t)$ controls the margin magnitude at training step
$t$, $s(t)\in[0,1]$ is a prescribed schedule, and $\tau>0$ is the temperature.
CXR uses a static margin with $s(t)=1$, while CT gradually increases the
margin according to the schedule specified in the experimental setup.
The indicator restricts the correction to unmatched pairs.

For either retrieval direction $q$, the resulting contrastive loss is
\begin{equation}
    \mathcal{L}_{q}
    = -\frac{1}{N}\sum_{i=1}^{N}
      \log
      \frac{\exp(\widetilde{S}^{q}_{ii})}
      {\sum_{j=1}^{N}\exp(\widetilde{S}^{q}_{ij})}.
    \label{eq:directional_loss}
\end{equation}
The metadata-conditioned contrastive term averages both directions:
\begin{equation}
    \mathcal{L}_{\mathrm{Meta}}
    = \frac{1}{2}\left(
      \mathcal{L}_{\mathrm{I2T}}+\mathcal{L}_{\mathrm{T2I}}
      \right).
    \label{eq:meta_clip_loss}
\end{equation}

Substituting Eq.~\eqref{eq:metadata_relation} into
Eq.~\eqref{eq:meta_logits} shows that each negative logit increases by
$\mu_t\alpha K^{\mathrm{meta}}_{ij}/\tau$. Acquisition-similar pairs therefore
contribute more strongly to the softmax normalizer. This relation-aware
hard-negative reweighting emphasizes distinctions among studies with similar
imaging conditions, while the positive logits retain their standard form.

\subsection{Training and Inference}
The acquisition relation is computed during pretraining and detached before
logit modulation. Categorical embeddings, bandwidths, field weights, and
$\alpha$ remain fixed during image--text optimization. The contrastive loss
therefore updates the encoders against a prescribed acquisition prior.
Inference uses the trained image and text encoders without DICOM metadata.

For CXR, Eq.~\eqref{eq:meta_clip_loss} is incorporated into the KAD training
objective. For CT, we combine it with the symmetric InfoNCE base objective
used by CT-CLIP:
\begin{equation}
    \mathcal{L}_{\mathrm{CT}}
    =\mathcal{L}_{\mathrm{InfoNCE}}+\gamma\mathcal{L}_{\mathrm{Meta}},
    \label{eq:ct_objective}
\end{equation}
where $\gamma$ controls the weight of the metadata-conditioned term. The
acquisition relation is shared across modalities, while each backbone retains
its base training objective.
\section{Experimental Setup}
We evaluate Meta-CLIP on 2D chest radiography and 3D chest CT, combining
zero-shot classification with analyses of representation structure, grounding,
and robustness to appearance perturbations.

\subsection{Backbones and Evaluation Protocol}
For CT, Meta-CLIP uses CT-ViT~\cite{hamamci2024generatect} and
CXR-BERT~\cite{boecking2022making}, with both encoders optimized following
CT-CLIP. Pretraining uses CT-RATE, whose official validation split provides
internal evaluation, while RAD-ChestCT provides external evaluation.
For CXR, we follow KAD~\cite{zhang2023knowledge}, training an M3AE-initialized
ViT-B/16 while keeping the pretrained BioClinicalMPBERT text
encoder\footnote{\url{https://huggingface.co/Laihaoran/BioClinicalMPBERT}}
frozen. The CXR benchmark comprises ChestXray14, CheXpert, PadChest-seen,
RSNA Pneumonia, and SIIM-ACR. Each task uses its designated evaluation split.

CXR pretraining uses 224$\times$224 inputs, batch size
32, 100 epochs, AdamW with learning rate $5\times10^{-5}$ and weight decay
0.02, cosine decay to $10^{-6}$, and 20 warm-up epochs. Its base contrastive
temperature is 0.07; the metadata margin uses $\mu_{\max}=0.5$, margin
temperature $\tau=0.1$, a static schedule, Gaussian bandwidth $\sigma=1.0$,
and $\alpha=0.5$.

The CT configuration uses batch size 8, learning rate
$5\times10^{-5}$, cosine decay with a $1.25\times10^{-6}$ minimum learning
rate, and 1,000 warm-up steps. The metadata term uses $\gamma=1.0$,
$\mu_{\max}=0.5$, and $\tau=0.1$. The acquisition kernel uses
$\sigma_d=1.0$, uniform field weights, and $\alpha=0.1$. The margin scale
$s(t)$ takes values $0.0$, $0.1$, $0.4$, $0.6$, and $0.8$ over epochs
$0$--$5$, $5$--$10$, $10$--$15$, $15$--$20$, and after epoch $20$,
respectively. Kernel parameters remain fixed in both modalities. CXR
incorporates the margin within the KAD objective and uses no separate
$\gamma$ coefficient.

We explore backbone portability by adding the metadata-conditioned loss
(weight $0.1$) to BrgSA~\cite{lai2025brgsa} with a five-field
Gaussian kernel. Both variants are trained on CT-RATE and selected by
pretraining validation loss, with evaluation on CT-RATE, RAD-ChestCT, and
INSPECT. INSPECT volumes use canonical LPS orientation. The metadata variant
uses a different batch size and omits BrgSA's local-alignment term, making this
an exploratory comparison of training configurations.

Zero-shot evaluation uses the pretrained model without target-dataset
parameter updates. Target evaluation labels and metadata are excluded from
adaptation.

\subsection{Metadata Selection and Characterization}
For CT-RATE, we use seven DICOM attributes describing scanner identity or
image formation: \texttt{XRayTubeCurrent}, \texttt{Exposure},
\texttt{ExposureTime}, \texttt{DistanceSourceToDetector}, \texttt{CTDIvol},
\texttt{ReconstructionDiameter}, and \texttt{Manufacturer}. This compact set
focuses on acquisition conditions. Demographic and unrestricted DICOM fields
can also encode disease prevalence or institutional label distributions.
We characterize the selected fields on the CT-RATE training set
($n=47{,}149$) using class-balanced logistic probes with an 80/20 stratified
split. Acquisition-domain signal is measured by balanced-accuracy lift over
chance when predicting scanner manufacturer, scanner model, and convolution
kernel. Disease association is measured by AUROC across the 18 CT-RATE
findings. These probes characterize the metadata independently of model
training.

CXR pretraining uses MIMIC-CXR-JPG image--report pairs and acquisition
metadata. Its kernel combines five numerical acquisition
fields (\texttt{KVP}, \texttt{ExposureIndex}, \texttt{ExposureTimeInuS},
\texttt{XRayTubeCurrent}, and \texttt{RelativeXRayExposure}) with the
categorical field \texttt{Manufacturer}. Training metadata comes exclusively
from the pretraining source. CheXpert, PadChest, ChestXray14, RSNA, and
SIIM-ACR supply no training metadata.

The CT field-set ablation compares the default set with three alternatives.
Physics-5 removes \texttt{ExposureTime} and \texttt{Manufacturer}.
Alternative-5 uses \texttt{ZSpacing}, \texttt{ExposureTime},
\texttt{NumberofSlices}, \texttt{RotationDirection}, and
\texttt{Manufacturer}. The final variant adds \texttt{XYSpacing} to the
default set. All variants share the CT-CLIP backbone, the objective in
Eq.~\eqref{eq:ct_objective}, a fixed acquisition relation, a frozen text
encoder, a cosine learning-rate schedule, and model selection by validation
AUROC. Freezing the text encoder isolates the field-set comparison from the
joint encoder fine-tuning used in the main CT model.

Table~\ref{tab:metadata_portability} summarizes acquisition fields exposed by
MIMIC-CXR-JPG and four public CXR resources. Availability refers to dataset
schemas; completeness can vary across individual studies. This variation
motivates selecting training fields according to source coverage, while
metadata-free inference accommodates evaluation datasets with different
schemas.

\begin{table}[!t]
\centering
\caption{Availability of representative DICOM fields across public CXR datasets.}
\label{tab:metadata_portability}
\scriptsize
\setlength{\tabcolsep}{3.0pt}
\renewcommand{\arraystretch}{1.08}
\begin{tabular}{lp{0.69\columnwidth}}
\toprule
\textbf{Dataset} & \textbf{Available fields} \\
\midrule
MIMIC-CXR-JPG & Mfr., KVP, ExpIndex, ExpTime, TubeCurrent, RelExp \\
CheXpert+ & Mfr., View, KVP, Exp., ExpTime, TubeCurrent, RelExp, PixelSpace, PhotoInterp \\
VinDr     & PhotoInterp \\
RSNA      & View, PixelSpace, PhotoInterp \\
PadChest  & Mfr., View, Exp., ExpTime, TubeCurrent, RelExp, PhotoInterp \\
\bottomrule
\end{tabular}
\end{table}

For CT robustness evaluation, we independently apply three MONAI intensity
transforms: \texttt{RandHistogramShift} with 10 control points,
\texttt{RandAdjustContrast} with $\gamma\in[0.85,1.15]$, and
\texttt{RandKSpaceSpikeNoise} with default intensity and channel-wise
injection. CXR evaluation tests metal artifacts, brightness shifts, and noise.
For retrieval, uniform sampling from the database provides a random reference.

\subsection{Ablations and Statistical Analysis}
The focused CXR comparison summarizes KAD and Meta-CLIP performance on
ChestXray14 and CheXpert (Table~\ref{tab:cxr_kad_comparison}), using the
corresponding results from the five-dataset benchmark in
Table~\ref{tab:seen}.

For acquisition-stratified analysis, CT-CLIP and Meta-CLIP predictions are
matched to the same cases and 18 target findings on CT-RATE. Continuous
attributes define quartile-based subgroups where ties permit. Within each
subgroup, AUROC is macro-averaged over findings containing both positive and
negative cases. We estimate 95\% confidence intervals using 10{,}000
case-bootstrap resamples, with paired resampling for model comparisons.
These intervals describe case-level uncertainty for the trained models and
exclude variability from retraining.

\begin{table}[!t]
\centering
\caption{Zero-shot CXR classification compared with KAD. Values are macro-AUROC (\%).}
\label{tab:cxr_kad_comparison}
\setlength{\tabcolsep}{4.0pt}
\renewcommand{\arraystretch}{1.08}
\begin{tabular}{lcc}
\toprule
\textbf{Method} & \textbf{ChestXray14} & \textbf{CheXpert} \\
\midrule
KAD & 76.49 & 89.23 \\
\textbf{Meta-CLIP (Ours)} & \textbf{80.65} & \textbf{91.31} \\
\bottomrule
\end{tabular}
\end{table}

\section{Results}
\noindent(\textbf{\romannumeral 1}) \textbf{\textit{Zero-shot disease classification.}}
Meta-CLIP improves CT disease classification on both internal and external
evaluation cohorts. On CT-RATE, macro-AUROC increases from 73.08\% to 74.49\%,
a gain of 1.41 points (95\% paired bootstrap CI, 0.96--1.85).
Figure~\ref{fig:ct_results}a shows improvements in 13 of 18 CT-RATE findings
and all seven displayed RAD-ChestCT findings. Nodule AUROC increases by 3.67
points on CT-RATE and 7.82 points on RAD-ChestCT, while lung opacity improves
by 9.38 and 4.73 points, respectively. The external results extend the
classification gains beyond the pretraining dataset.

On CXR, Meta-CLIP achieves the highest macro-AUROC on ChestXray14,
PadChest-seen, RSNA Pneumonia, and SIIM-ACR among the methods in
Table~\ref{tab:seen}. The focused comparison with KAD in
Table~\ref{tab:cxr_kad_comparison} shows macro-AUROC gains of 4.16 points
on ChestXray14 (76.49\% to 80.65\%) and 2.08 points on CheXpert
(89.23\% to 91.31\%). These values summarize the overall model comparison
on the two benchmarks.

\begin{figure}[!t]
    \centering
    \includegraphics[width=0.88\columnwidth]{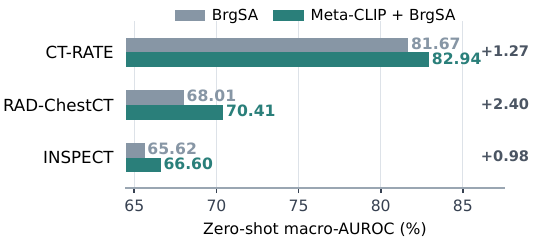}
    \caption{\textbf{CT backbone portability.} Zero-shot macro-AUROC of BrgSA and Meta-CLIP + BrgSA on CT-RATE, RAD-ChestCT, and INSPECT.}
    \label{fig:brgsa_portability}
\end{figure}

The exploratory BrgSA comparison in Fig.~\ref{fig:brgsa_portability} shows
macro-AUROC gains of 1.27 points on CT-RATE, 2.40 points on RAD-ChestCT, and
0.98 points on INSPECT. These results suggest that acquisition-aware training
can also benefit another volumetric backbone. As a complementary retrieval
analysis, Fig.~\ref{fig:ct_results}b reports top-1 label similarities of 0.958
on RAD-ChestCT and 0.886 on CT-RATE, using random retrieval as a reference.

\noindent(\textbf{\romannumeral 2}) \textbf{\textit{Representation structure.}}
Figure~\ref{fig:embedding_umap} examines how the classification gains relate
to embedding organization. On CT, disease-cluster silhouette scores increase
from 0.0437 to 0.1085 for lymphadenopathy and from 0.0243 to 0.0721 for lung
opacity. The dataset-distribution silhouette decreases from 0.1533 to 0.1206,
indicating less separation by dataset. On CXR, Meta-CLIP also yields a higher
disease-cluster silhouette than KAD. Together, these analyses suggest that the
learned geometry is more closely aligned with disease categories.

\begin{figure*}[!t]
    \centering
    \includegraphics[width=0.90\textwidth]{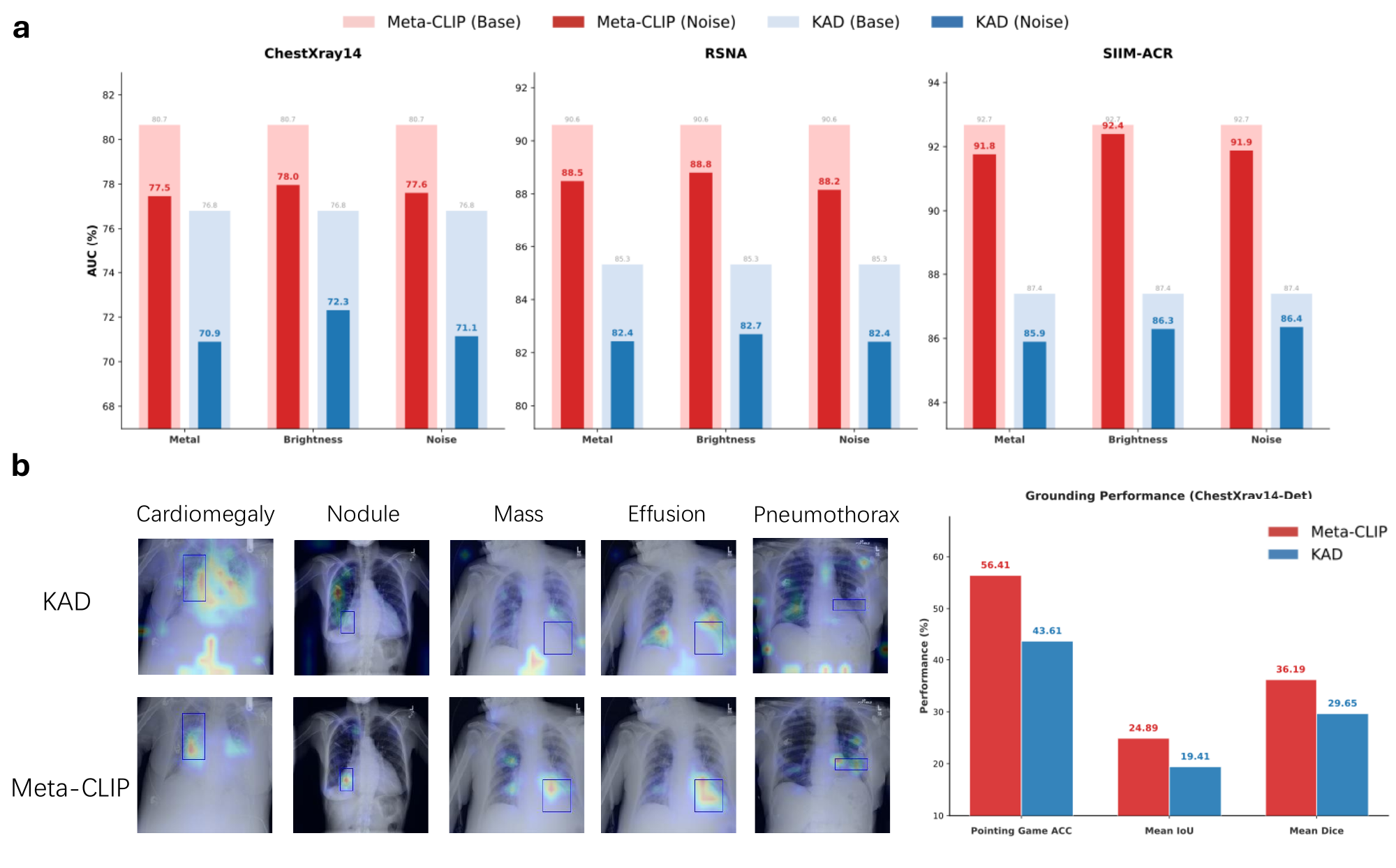}
    \caption{\textbf{Robustness and grounding study on CXR.} (a) Zero-shot AUC under metal artifacts, brightness shifts, and noise on ChestXray14, RSNA, and SIIM-ACR (light: clean; dark: perturbed). (b) ChestXray14-Det grounding: cross-attention maps for five findings (blue: ground-truth boxes) and metrics (Pointing Game, mIoU, and Dice).}
    \label{fig:xray_results}
\end{figure*}

\noindent(\textbf{\romannumeral 3}) \textbf{\textit{Disease grounding.}}
The grounding evaluation tests whether improved image--text alignment also
corresponds to better localization. On ChestXray14-Det,
Fig.~\ref{fig:xray_results}b shows gains over KAD of 12.80 points in Pointing
Game accuracy, 5.48 points in mean IoU, and 6.54 points in mean Dice. These
results indicate stronger spatial correspondence between textual findings
and image evidence.

\noindent(\textbf{\romannumeral 4}) \textbf{\textit{Robustness to appearance perturbations.}}
Figure~\ref{fig:ct_results}c compares CT performance under synthetic intensity
perturbations. Under histogram shift, macro-AUROC decreases by 0.41 points
for Meta-CLIP and 1.04 points for CT-CLIP. On CXR,
Fig.~\ref{fig:xray_results}a shows smaller degradation than KAD under metal
artifacts, brightness shifts, and noise across ChestXray14, RSNA, and
SIIM-ACR. These comparisons use the clean and perturbed inputs within each
robustness protocol; the five-dataset classification benchmark is reported
separately in Table~\ref{tab:seen}.

\noindent(\textbf{\romannumeral 5}) \textbf{\textit{Acquisition information in the selected metadata.}}
Table~\ref{tab:metadata_characterization} characterizes the information
carried by the selected CT fields. Coverage is nearly complete except for
source-to-detector distance, which is available for 47.9\% of scans.
The numerical kernel excludes unobserved fields from pairwise comparisons.
Individual fields predict acquisition-domain proxies while showing limited
disease association, with mean disease AUROC ranging from 0.502 to 0.542.
Jointly, the seven fields yield a mean balanced accuracy of 0.774 across
the three acquisition-domain proxies and a mean disease AUROC of 0.593.
Because manufacturer is included among the inputs, the manufacturer probe
describes information retained by the field set rather than independent
domain prediction. The unrestricted pool of 43 attributes yields higher
domain prediction (0.995) but also higher disease AUROC (0.724). This
comparison supports focusing on acquisition fields with lower direct disease
association.

\begin{table}[!t]
\centering
\caption{Characterization of selected CT metadata. Values report coverage, acquisition-domain lift, and disease AUROC.}
\label{tab:metadata_characterization}
\setlength{\tabcolsep}{2.6pt}
\renewcommand{\arraystretch}{1.08}
\begin{tabular}{lccc}
\toprule
\textbf{Metadata field} & \textbf{Coverage} & \textbf{Domain lift} & \textbf{Disease AUROC} \\
\midrule
Tube current              & 100.0\% & 0.131 & 0.536 \\
Exposure                  & 100.0\% & 0.126 & 0.536 \\
Exposure time             & 100.0\% & 0.331 & 0.502 \\
Source--detector distance & 47.9\%  & 0.351 & 0.524 \\
CTDIvol                   & 99.8\%  & 0.183 & 0.542 \\
Reconstruction diameter   & 100.0\% & 0.040 & 0.519 \\
Manufacturer              & 100.0\% & 0.345 & 0.533 \\
\bottomrule
\end{tabular}
\end{table}

\noindent(\textbf{\romannumeral 6}) \textbf{\textit{Effect of metadata selection.}}
Table~\ref{tab:metadata_set_ablation} tests whether additional or alternative
fields improve transfer. The default seven-field set gives the highest
macro-AUROC point estimates on CT-RATE and RAD-ChestCT. Adding
\texttt{XYSpacing} decreases performance by 0.63 and 0.58 points,
respectively. Alternative-5 retains similar external performance but loses
1.18 points on CT-RATE, while Physics-5 also scores below the default on
both datasets. The confidence intervals overlap, particularly on RAD-ChestCT,
so these comparisons support the compact default as a practical configuration
within the evaluated field sets.

\begin{table}[!t]
\centering
\caption{Effect of metadata selection on CT zero-shot classification. Values are macro-AUROC (\%) with 95\% bootstrap confidence intervals.}
\label{tab:metadata_set_ablation}
\setlength{\tabcolsep}{4.0pt}
\renewcommand{\arraystretch}{1.08}
\resizebox{\columnwidth}{!}{%
\begin{tabular}{lcc}
\toprule
\textbf{Metadata set} & \textbf{CT-RATE} & \textbf{RAD-ChestCT} \\
\midrule
Alternative-5              & 73.21 [72.55,73.89] & 65.45 [64.76,66.12] \\
Physics-5                   & 73.30 [72.65,73.97] & 65.33 [64.65,66.00] \\
Default + \texttt{XYSpacing} & 73.76 [73.11,74.43] & 64.92 [64.24,65.60] \\
Default compact-7           & \textbf{74.39 [73.75,75.04]} & \textbf{65.50 [64.82,66.17]} \\
\bottomrule
\end{tabular}
}
\end{table}

\noindent(\textbf{\romannumeral 7}) \textbf{\textit{Pairwise acquisition similarity.}}
We examine the acquisition kernel on 999{,}921 non-self pairs sampled from
12{,}000 CT-RATE training volumes. Aggregate similarity is
$0.657\pm0.216$ (median $0.688$), with an average of 5.23 jointly observed
numerical fields per pair (range: 4--6). Source-to-detector distance is jointly
observed in 23.5\% of pairs. The fields contribute different similarity
distributions: excluding \texttt{ExposureTime}, their means range from 0.597
to 0.684 and medians from 0.678 to 0.821. In contrast,
\texttt{ExposureTime} is concentrated near one (median $0.999$,
IQR $0.999$--$1.000$), contributing little pairwise variation at the chosen
bandwidth. The other numerical fields provide a broader range of similarities.
This diagnostic covers numerical fields; \texttt{Manufacturer} enters the
categorical kernel.

\noindent(\textbf{\romannumeral 8}) \textbf{\textit{Performance across acquisition strata.}}
Table~\ref{tab:acquisition_subgroups} examines CT-RATE performance within
manufacturer and acquisition-parameter subgroups. Meta-CLIP improves
macro-AUROC for all three manufacturer groups, including gains of 2.02 points
for Siemens and 2.37 points for the smaller PNMS group. Gains also remain
in the highest CTDIvol, tube-current, exposure, and reconstruction-diameter
strata, where baseline performance is below its overall average. Across the
22 evaluated strata, 21 improve. The longest exposure-time stratum
($n=179$) decreases by 0.54 points. These descriptive results indicate that
the overall gain extends across multiple acquisition settings, with
heterogeneity in the size and direction of subgroup effects.

\begin{table}[!t]
\centering
\caption{Acquisition-stratified CT-RATE validation results. Values are macro-AUROC (\%) over evaluable findings.}
\label{tab:acquisition_subgroups}
\scriptsize
\setlength{\tabcolsep}{3.2pt}
\renewcommand{\arraystretch}{1.05}
\begin{tabular}{lrrrr}
\toprule
\textbf{Subgroup} & \textbf{$n$} & \textbf{CT-CLIP} & \textbf{Meta-CLIP} & \textbf{$\Delta$} \\
\midrule
All cases                 & 3039 & 73.08 & 74.49 & +1.41 \\
Philips                   & 1928 & 74.06 & 75.04 & +0.97 \\
Siemens                   &  971 & 70.65 & 72.67 & +2.02 \\
PNMS                      &  140 & 73.85 & 76.21 & +2.37 \\
Highest CTDIvol           &  759 & 71.01 & 72.51 & +1.50 \\
Highest tube current      &  756 & 72.78 & 74.68 & +1.90 \\
Highest exposure          &  758 & 71.39 & 72.68 & +1.29 \\
Largest recon. diameter   &  758 & 69.05 & 71.05 & +2.00 \\
Longest exposure time     &  179 & 69.46 & 68.92 & $-$0.54 \\
\bottomrule
\end{tabular}
\end{table}

\begin{table*}[!t]
\centering
\caption{Zero-shot disease classification across five CXR datasets for findings represented during pretraining. Values are point estimates (\%). Best results are bolded and second-best results are underlined.}
\label{tab:seen}
\resizebox{\textwidth}{!}{
\begin{tabular}{l|ccc|ccc|ccc|ccc|ccc}
\toprule
\textbf{Dataset} & \multicolumn{3}{c|}{\textbf{ChestXray-14}} & \multicolumn{3}{c|}{\textbf{CheXpert}} & \multicolumn{3}{c|}{\textbf{PadChest-seen}} & \multicolumn{3}{c|}{\textbf{RSNA Pneumonia}} & \multicolumn{3}{c}{\textbf{SIIM-ACR}} \\
\textbf{Method} & AUC$\uparrow$ & F1$\uparrow$ & ACC$\uparrow$ & AUC$\uparrow$ & F1$\uparrow$ & ACC$\uparrow$ & AUC$\uparrow$ & F1$\uparrow$ & ACC$\uparrow$ & AUC$\uparrow$ & F1$\uparrow$ & ACC$\uparrow$ & AUC$\uparrow$ & F1$\uparrow$ & ACC$\uparrow$ \\
\midrule

ConVIRT ~\cite{zhang2022contrastive}
& 53.15 & 12.38 & 57.88 
& 52.10 & 35.61 & 57.43 
& 63.72 & 14.56 & 73.47 
& 79.21 & 55.67 & 75.08 
& 64.25 & 42.87 & 53.42 \\

GLoRIA  ~\cite{huang2021gloria}    
& 55.92 & 14.20 & 59.47 
& 54.84 & 37.86 & 60.70 
& 64.09 & 14.83 & 73.86 
& 70.37 & 48.19 & 70.54 
& 54.71 & 40.39 & 47.15 \\

BioViL   ~\cite{boecking2022making}
& 57.82 & 15.64 & 61.33 
& 60.01 & 42.10 & 66.13 
& 60.35 & 10.63 & 70.48 
& 84.12 & 54.59 & 74.43 
& 70.28 & 46.45 & 68.22 \\

BioViL-T ~\cite{boecking2022making}
& 60.43 & 17.29 & 62.12 
& 70.93 & 47.21 & 69.96 
& 65.78 & 15.37 & 77.52 
& 86.03 & 62.56 & 80.04 
& 75.56 & 60.18 & 73.72 \\

CheXzero ~\cite{tiu2022expert}     
& 66.99 & 19.95 & 65.38 
& 87.90 & 61.90 & 81.17 
& 73.24 & 19.53 & 83.49 
& 83.13 & 61.49 & 78.34 
& 84.60 & 65.97 & 77.34 \\

MedKLIP~\cite{wu2023medklip} 
& 72.33 & 24.18 & 79.40 
& 87.97 & 63.67 & 84.32 
& 77.87 & 26.63 & 92.44 
& 86.57 & 63.28 & 79.97 
& 89.79 & 72.73 & 83.99 \\

KAD ~\cite{zhang2023knowledge}
& 76.49 & 29.98 & 80.49 
& 89.23 & 63.25 & 86.25 
& 80.94 & 58.23 & 92.12 
& 85.32 & 87.09 & 81.80 
& 87.39 & 68.84 & 81.73 \\

MAVL ~\cite{phan2024decomposing}
& 73.57 & 26.25 & \textbf{82.77}
& 90.13 & \textbf{65.47} & 86.44
& 78.79 & 28.48 & \underline{92.56} 
& \underline{86.91} & 63.41 & \underline{82.42} 
& 92.04 & \underline{77.95} & \underline{87.14} \\

CARZero ~\cite{lai2024carzero}
& \underline{79.64} & \underline{31.41} & 79.60 
& \textbf{92.38} & 40.20 & 86.93 
& 83.97 & 25.08 & 91.46 
& 80.28 & 28.28 & 41.52 
& \underline{90.90} & 75.96 & 85.03 \\

DeViDe ~\cite{luo2025devide}
& 77.61 & 31.40 & 82.06 
& 89.87 & 62.88 & \underline{87.98} 
& \underline{84.26} & \underline{58.29} & \underline{92.16} 
& 88.58 & \underline{88.40} & 84.00 
& 89.50 & 72.24 & 83.72 \\


\hline
\textbf{Meta-CLIP (Ours)} 
& \textbf{80.65} & \textbf{35.26} & \underline{82.13}
& \underline{91.31} & \underline{64.85} & \textbf{88.86}
& \textbf{86.00} & \textbf{60.25} & \textbf{93.32} 
& \textbf{90.60} & \textbf{90.23} & \textbf{86.42} 
& \textbf{92.68} & \textbf{78.62} & \underline{87.11} \\

\bottomrule
\end{tabular}}
\end{table*}

\section{Discussion}
\subsection{Acquisition Metadata as a Training-Time Prior}
Acquisition metadata provides context for contrastive comparisons beyond the
image and report embeddings. Meta-CLIP uses this context to emphasize
differences among studies acquired under similar conditions. The CXR benchmark
shows improvements over KAD, while the CT evaluation demonstrates gains on internal and
external cohorts. The BrgSA comparison provides exploratory evidence for
application to another volumetric backbone. Together, these findings support
acquisition similarity as a useful prior for weighting negative pairs.
The mechanism acts on negative comparisons and does not explicitly align
same-disease studies across acquisition conditions; conditional acquisition
invariance remains a separate property to evaluate.

\subsection{Practical Implications}
Meta-CLIP retains the inference computation, memory footprint, and input
requirements of its underlying dual encoder. A deployment site therefore
requires images and text prompts, with no additional DICOM input. This
separation is useful when metadata availability differs across institutions,
as illustrated by the CXR field inventory. During pretraining, field selection
should consider coverage and information content together. The compact CT
set carries less direct disease signal than the unrestricted pool, while
adding \texttt{XYSpacing} fails to improve the evaluated transfer results.
These observations favor source-specific selection of complementary
acquisition fields over indiscriminate inclusion of available metadata.

\subsection{Limitations}
The study uses retrospective public chest CT and CXR cohorts. Prospective,
multi-institutional evaluation is needed to assess transfer in clinical
workflows. The synthetic perturbations test selected appearance changes and
cover only part of the variation introduced by real scanners and workflows.
In addition, DICOM fields can be missing, inconsistently encoded, or correlated
with institution and patient population. Evaluation under unseen acquisition
protocols and heterogeneous metadata schemas will further characterize the
applicability of the acquisition prior.

\balance
\section{Conclusion}
We introduced Meta-CLIP, which incorporates acquisition context into medical
vision--language pretraining through pairwise contrastive margins. Its kernel
combines numerical and categorical DICOM fields to emphasize negative
comparisons between acquisition-similar studies. This fixed prior shapes
optimization while preserving the image and text encoders and their standard
inference procedure. Metadata is used during pretraining, with no additional
inputs at deployment.

Experiments on chest radiography and volumetric CT show improved zero-shot
classification, smaller performance losses under appearance perturbations,
and stronger disease grounding. The CXR benchmark demonstrates gains over
KAD, while external CT evaluation extends the transfer evidence beyond the
pretraining cohort. Representation analyses and metadata characterization
provide complementary views of the learned embeddings and their acquisition
prior. Together, these findings support acquisition context as a practical
complement to image--report supervision for learning transferable medical
representations.





\clearpage
\nobalance
\IEEEtriggeratref{12}
\bibliographystyle{IEEEtran}
\bibliography{reference}

\end{document}